\documentclass[11pt,a4paper]{article}
\usepackage{emnlp2022}
\usepackage{times}
\usepackage{latexsym}
\usepackage{tabularx}
\usepackage{graphicx}
\usepackage{float}
\usepackage{enumerate}
\usepackage{amssymb,amsmath}
\usepackage{color}
\usepackage{bm}
\usepackage{makecell}
\usepackage{multi row}
\usepackage{array}
\usepackage{resizegather}
\usepackage{refcount}
\usepackage{fixfoot}
\usepackage{hyperref}
\usepackage{url}
\usepackage{subcaption}  
\usepackage{bbm}

\usepackage{algorithm}
\usepackage{algorithmicx}
\usepackage{algpseudocode}

\newif\ifcomment
\commenttrue
\ifcomment

\else
\fi

\newcommand\tstrut{\rule{0pt}{2.6ex}}         
\newcommand\bstrut{\rule[-1.0ex]{0pt}{0pt}}   
\newcommand{\thinline}{\Xhline{1.5\arrayrulewidth}}
\newcommand{\thickline}{\Xhline{2.5\arrayrulewidth}}
\newcommand{\tsep}	{\bstrut \\ \thinline}

\newcommand{\ttop}{\thickline}
\newcommand{\tbottom}{\bstrut \\ \thickline}

\newcommand{\aln}[1] {
	\begin{align} #1 \end{align}
}
\newcommand{\alns}[1] {
	\begin{align*} #1 \end{align*}
}

\newcommand{\gradt}[1]{\nabla_{#1} \mathcal{L}_{t}}
\newcommand{\gradi}[1]{\nabla_{#1} \mathcal{L}_{i}}

\title{Meta-Learning Fast Weight Language Models}

\author{
Kevin Clark \qquad Kelvin Guu \qquad Ming-Wei Chang \\
\textbf{Panupong Pasupat \quad Geoffrey Hinton \quad Mohammad Norouzi} \\[.5em]
Google Research \\
{\fontsize{10}{11}\selectfont \texttt{\{kevclark,kguu,mingweichang,ppasupat,geoffhinton,mnorouzi\}@google.com}}
}
 
\date{}

\begin{document}
\maketitle

% Fast weight methods 
% Add meta-learning somewhere
% Go back to dynamic evaluation

\begin{abstract}
%Fast weight methods where the model parameters are updated during inference 
%have proven successful at few-shot adaptation.
%However, they are often computationally expensive to use in practice.
%We study 
% ---
%For instance, in the language modeling domain, dynamic evaluation updates model parameters at test time using gradient information from previous tokens. 
%While dynamic evaluation substantially improves language modeling performance, it requires over 3x the compute of standard inference. 

%\kc{I went back to a less exciting but less over-claiming abstract, as the method makes dynamic eval efficient, not FW in general.}
%\mw{I in fact like this version}
Dynamic evaluation of language models (LMs) adapts model parameters at test time using gradient information from previous tokens and
substantially improves LM performance. However, it requires over 3x more compute than standard inference. 
We present Fast Weight Layers (FWLs), a neural component that provides the benefits of dynamic evaluation much more efficiently by expressing gradient updates as linear attention.  
A key improvement over dynamic evaluation %\mw{maybe: "FWL can be trained to make god use ....}
is that FWLs can also be applied at training time so the model learns to make good use of gradient updates. %\mw{Do you want to make a coonection to meta learning here?}
FWLs can easily be added on top of existing transformer models, require relatively little extra compute or memory to run, and significantly improve language modeling perplexity. %\mw{Consider saying that you save the speed of dynamic evaluation 6 times instead of perplexity improvement.}
\end{abstract}

\section{Introduction}
A key challenge in language modeling is representing the contextual information from previous tokens. 
Transformer language models use attention to pass along this information, but constantly referring back to the previous text is a cognitively implausible model of working memory. 
%is very different from how humans read. \mw{I felt that while this is true, but it is probably not the main issue. Isn't the main issue is that the parameters are fixed during inference time? Should we hightlight that and then say fast weight allows parameters to change on the fly?}
An appealing alternative is using fast weight neural networks \citep{hinton1987using, schmidhuber1992learning}.
Inspired by short-term plasticity in the brain, these models have parameters that change on-the-fly based on input data (previous tokens for LMs) in addition to standard ``slow" weights learned during training.
Fast weights have proven successful for supervised \citep{ba2016using}, reinforcement \citep{munkhdalai2019metalearned}, and few-shot \citep{munkhdalai2017meta} learning.

Dynamic evaluation \citep{mikolov2010recurrent,krause2018dynamic} uses a variant of fast weights to improve language models at inference time.
After scoring (or generating) a chunk of text, dynamic evaluation applies a gradient update to the model coming from the LM loss over that chunk before continuing.
Intuitively, this process improves performance because an update that makes the model better at predicting previous tokens will likely also make it better at predicting future ones. 
Dynamic evaluation substantially improves LM perplexity, but has numerous drawbacks.
It requires an extra forward and backward pass through the model to compute the gradients, and the sequential gradient updates over chunks cannot be parallelized.
%Furthermore, a separate copy of the evolving model weights has to be stored for each example in a minibatch, so, in practice, implementations use a batch size of 1 to reduce memory usage.
Furthermore, it is very memory intensive because a separate copy of the evolving model weights has to be stored for each example in a minibatch.
Lastly, dynamic evaluation is only used at test-time, so the model does not learn to make good use of gradient updates.

% \mw{I prefer using bullet points over advantages of our methods rather than disadvantages of other methods. Removing bullet points here also might save you some space.}
% \vspace{-1mm}
% \begin{itemize}
%     \item An extra forward and backward pass is required to compute the gradients. \vspace{-1mm}
%     \item The sequential gradient updates on different text chunks cannot be parallelized. \vspace{-1mm}
%     \item A separate copy of the evolving model weights has to be stored for each example in a minibatch. So in practice implementations use batch size 1 to reduce memory usage. \vspace{-1mm}
%     \item Dynamic evaluation is only used at test-time. The model does not learn to make good use of gradient updates, and the update rule has to be hand-tuned rather than learned. \vspace{-1mm}
% \end{itemize}

\begin{figure*}[tb!]
\begin{center}
\includegraphics[width=0.95\textwidth]{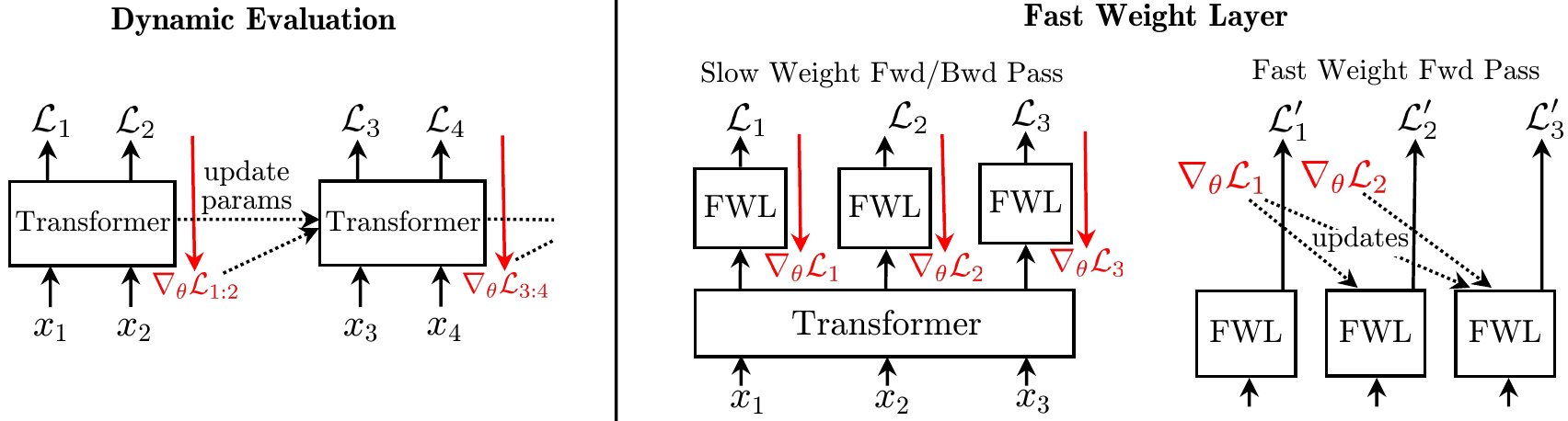}
\end{center}
\vspace{-1mm}
\caption{Dynamic evaluation (left) recurrently processes chunks of text and applies the gradient updates to model weights.
FWLs (right) compute the gradients in parallel and for only a small subset of model parameters, greatly increasing efficiency. The improved speed allows FWLs to be used at train time, making them more effective.}
\vspace{-1mm}
\label{fig:overview}
\end{figure*}

We present Fast Weight Layers\footnote{Code is at \url{github.com/google-research/google-research/tree/master/fwl}.} (FWLs), a neural component that provides the benefits of dynamic evaluation with none of these downsides.
They can be added to any LM without requiring changes to the training or evaluation loop.
Like dynamic evaluation, FWLs also update their parameters using gradient information from previous tokens, but FWLs employ three key ideas to improve efficiency.
First, FWLs are added on top of the transformer after the last attention layer, which avoids having to backpropagate through the whole transformer and circumvents the complexities of backpropagation through time. 
Secondly, FWLs compute gradients in parallel rather than recurrently.
Lastly, FWLs leverage the property that gradient matrices are rank one to compute their outputs efficiently. 

Crucially, the efficient design means FWLs can be used at training time so the model learns to obtain beneficial updates from previous gradients.
Another benefit is that while dynamic evaluation needs hyperparameter search to find a good step size for the gradient update, FWLs can instead learn the step size. 
Training FWLs can be viewed as applying gradient-based meta-learning \citep{finn2017model} to language modeling, where the support set contains tokens seen so far and the query set contains future tokens, a perspective which helps explain some of the behaviors of FWLs.

% Talk about what the baselines are
% How FWLs can easily be added
% Use as opportunity to cite more prior work

%FWLs can be easily added to existing text generation architectures.
FWLs scale well to long sequences and are complementary to existing long-text generation methods such as sparse attention \citep{child2019generating} or recurrent processing \citep{dai2019transformer}.
We evaluate FWLs at language modeling on the WikiText-103 dataset \citep{merity2016pointer}.
They substantially improve perplexities over strong baselines; for example lowering Transformer-XL's perplexity from 18.1 to 16.6.
This gain is comparable to the one from dynamic evaluation, but FWLs add less than 20\% compute overhead to the model compared to the over 200\% of dynamic evaluation.
 %, which we showcase by adding them to sparse tranformer models as well as the long-context Transformer-XL model \citep{dai2019transformer}.
Ablations show FWLs achieve superior compute vs perplexity trade-offs compared to alternative fast weight methods. We also analyze how FWLs improve results and find they are especially effective at modeling rare tokens, repeated tokens, and long documents.

\section{Method}

%\mw{I would prefer some high level points first. For example, say there are two types of parameters the slow one and the fast one, and we are going to on-the-fly update the fast one while keep the slow one untouched ,.....}

Our models first run a transformer decoder (or other left-to-right neural network) over an input sequence $[x_1, ..., x_T]$, yielding vector representations $[h_1, ..., h_T]$. %\mw{Stupid question: it seems that there are also the original transformer parameter. That is different from the slow weight $\theta$? is it?}
FWLs use a learned function $f_\theta$ with parameters $\theta$ to further process the text.
These parameters (and the transformer's parameters) are ``slow" weights learned during training.
We chose to use a hidden layer, projection layer, and LayerNorm \citep{Ba2016LayerN} for $f_\theta$, which worked well in initial experiments:
\alns{
f_\theta(h_t) = \text{LayerNorm}(\text{ReLU}^2(h_t U + a)W + b)
}
The FWL's output can be passed into a prediction layer such as an output softmax:
\alns{
  \mathcal{L}_t &= \text{CE}(\text{softmax}(f_\theta(h_t)E + c), x_{t+1})   
}
where $E$ is an embedding matrix, $c$ is a bias vector, and CE denotes cross-entropy loss. 
FWLs also work with adaptive softmax layers \citep{Grave2017EfficientSA}, such as the one used by Transformer-XL.
%For Transformer-XL, we instead use an adaptive softmax \.

In addition to the slow weights, an FWL also employs ``fast'' weights that let it adapt to contextual information. 
First, the FWL does a forward and backward pass over the input sequence with its slow weights $\theta$ to obtain a gradient $\gradt{\theta}$ from each position $t$.
Importantly, these $T$ gradients can be computed in a single backwards pass because $\mathcal{L}_t$ depends only on $h_t$. 
If we added attention or other temporal dependence to the FWL, backpropagation through time would instead only compute the summed gradient $\sum_{t=1}^T \gradt{\theta}$. This limitation is why dynamic evaluation has to process the input in chunks, which means the model does not get updates from recent tokens within the same chunk; FWLs do not have this drawback. 

%(If $L_t$ depended on previous $h$'s as well, e.g.\ if we added attention to the FWL, backpropagation through time could instead only compute the summed gradient $\sum_{t=1}^T \gradt{\theta}$.)

The FWL then does a second forward pass using evolving fast weights $\theta'$.
The fast weights are initialized with the slow weights $\theta$.
Conceptually, the second pass processes the sequence in a left-to-right order, although we will show this can be parallelized in practice.\footnote{Generating with a LM is inherently recurrent, but FWLs avoid recurrence when training the model or scoring text.}
For each position $t$, it re-runs the FWL and output softmax on $h_t$, but now using $\theta'$ instead of $\theta$ to produce a new loss $\mathcal{L}'_t$.
%During training or perplexity evaluation it outputs a new loss $\mathcal{L}'_t$; during generation it instead outputs the next token.
Then the FWL updates its fast weights as $\theta' \leftarrow \theta' - \alpha \circ \gradt{\theta}$ where $\alpha$ consists of learned step sizes. We learn a separate step size for each weight matrix/vector in the FWL, which performs slightly better than having one global step size. 
%where $\alpha$ is a learned step size, adapting the model to the current token.
%Note this gradient update is using the slow weight gradient from $\mathcal{L}_t$ rather than from $\mathcal{L}'_t$ because $\nabla_\theta \mathcal{L}_t$ can be computed in parallel. 
Although we tried more sophisticated update rules such as one based on Adam, they did not outperform this simpler one in initial experiments. An overview of FWLs is shown in Figure~\ref{fig:overview}.
%and how they compare to dynamic evaluation 
%Generating 

%The final losses $\mathcal{L}_t'$ are used for training the model or evaluating its performance at test time.

%\vspace{1mm}
\paragraph{Efficient Fast Weight Computation.}
We will now show how the FWL outputs can be computed without needing to store the $T$ gradients of $\mathcal{L}_{1:T}$ in memory (which would have prohibitive memory requirements) or recurrently computing $\theta'$ (which would be slow on modern accelerators).
We will illustrate this for the second matrix-multiply in $f_\theta$; other parts of the FWL can be computed analogously. 
We use $v_t$ to denote the input, $o_t$ to denote the output, and the $'$ symbol to differentiate activations in the second (fast weight) forward pass from those in the first (slow weight) pass.

The slow weight output at position $t$ is simply
$o_t = v_t W$.
The fast weight output at position $t$ is
\aln{
    o'_t &= v'_t (W - \alpha_{W}\sum_{i<t} \gradi{W}) \label{eqn:1}
}
where the weight matrix $W$ has been updated by the $t - 1$ previous gradients.

Computing (\ref{eqn:1}) naively is infeasible because it requires storing all $T$ gradient matrices in memory.
%Computing the outputs in this way has runtime $\mathcal{O}(nmT)$, which is desirable because it scales well with sequence length. 
%In practice however, it is infeasible to use because (1) it requires storing all $T$ $n \times m$ gradient matrices in memory and (2) the partial sums requires sequential computation.
% The full $T \times m$ matrix of outputs $O$ is
% \alns{
%   O' = V' \text{cumsum}\left(\grad{W}\right)
% }
% where $V \in \mathbb{R}^{T \times n}$ and $O \in \mathbb{R}^{T \times m}$ are matrices of activations. This method takes $\mathcal{O}(nmT)$ compute, but in practice it is infeasible to use because (1) it requires storing all $T$ $n \times m$ gradient matrices in memory and (2) the cumsum requires sequential computation.
%Let $\delta_i$ be the upstream gradient $\gradi{o_i}$.
As an alternative, we develop a more efficient method by taking advantage of the observation that
the gradient is equal to the outer product of the input and the upstream gradient: $\gradi{W} = v_i^T \gradi{o_i}$ (from the chain rule; see for example Appendix D of \citet{Mitchell2022FastME} a derivation). Therefore we can rewrite (\ref{eqn:1}) as
\aln{
    o'_t &= v'_t W - \alpha_{W}\sum_{i<t} v'_t v_i^T \gradi{o_i} \label{eqn:2}
}
The first term is just a regular matrix-multiply and can be computed easily. 
Interestingly, the sum in the second term can be interpreted as linear attention (i.e., without a softmax) using query $v'_t$, keys $v_i$, and values $\gradi{o_i}$. The partial sum up to $t$ corresponds to causal masking in the attention. 

Computing the FWL output using (\ref{eqn:2}) means we only need to store gradient vectors $\gradi{o_i}$ in memory (which is no worse than storing model activations) rather than gradient matrices $\gradi{W}$. It also can be computed fully in parallel using causal attention.
However, naive attention scales quadratically with sequence length, which is an obstacle for applying (\ref{eqn:2}) to long documents.
Luckily, because the attention is linear, there exist more efficient methods for computing it.
In particular, we use the mixed chunk attention method from \citet{hua2022transformer}, which improves efficiency through breaking the document into chunks  while still computing the attention scores exactly.

% Therefore the full $T \times m$ matrix of outputs $O$ can be written as
% \aln{
%   O' = V' W - \alpha (V' V^T \circ M) \grad{O} 
% }
% where where $V \in \mathbb{R}^{T \times n}$ and $O \in \mathbb{R}^{T \times m}$ are matrices of activations and $M \in \mathbb{R}^{T \times T}$ is a lower triangular mask enforcing causal masking.
% Naively computing $O'$ in this way has runtime $\mathcal{O}(T^2(m + n))$.
% However, we are interested in applying fast weights to long documents where $T >> n,m$. 
% As a solution, we use the mixed chunk attention from \citet{hua2022transformer}. 

% Mixed-chunk attention essentially offers a middle ground between (1) which has a desirable runtime but poor parallelization and (2) which is fully parallizable but has a $O(T^2)$ runtime. 
% The key ideas is to break the input sequence into $C$ chunks and use (1) across chunks but (2) withing chunks. More formally, the FWL output can be written as.
% \alns{
%     o'_t = v'_t (W - \alpha(&\sum_{c=1}^{\lfloor t / c \rfloor} \nabla_{W} \mathcal{L}_{cT/C:(c+1)T/C} \\
%          + &\sum_{\lfloor t / c \rfloor + 1}^{t - 1} v'_t v_i^T \gradi{o_i} ))
% }
% where the first sum is (1) applied to previous chunks and the second term is (2). This method requires storing $C$ gradient matrices, which in practice can be less memory than the neural network activations. 

Viewing the fast weight output as linear attention makes it clearer how FWLs work.
%The error signals $\gradi{o_i}$ show how to adjust the outputs $o_i$ such that the model will have lower loss, and the attention score $v'_t v_i^T$ measures the similarity between the current position and previous ones. The more similar the previous position, the more the FWL's output is adjusted in the direction of $\gradi{o_i}$.
For each previous position $i$, the error signal $\gradi{o_i}$ is the direction for adjusting the output $o_i$ to lower the loss, and the attention score $v'_t v_i^T$ measures the similarity between position $i$ and the current position. This means FWLs adjust the current output in a way that mimics the loss-reducing directions of similar past positions. %\ice{Edited this a bit. PTAL} \kc{LGTM!}
%Some fast weight methods have a decay parameter that lowers the contribution from further-away tokens, but we did not find this idea to improve results. The similarity score $v'_t v_i^T$ suggests why: the model can learn representations such that further-away tokens have lower similarity, avoiding the need to explicitly decay the fast weights over time. 

Computing fast weights for vectors (e.g.\ biases or LayerNorm weights) rather than matrices is simpler because the gradients require less memory. For example, the fast weight output after adding the bias term $b$ to $o_t$ is $o'_t + b - \alpha_{b}\sum_{i<t} \gradi{b}$. This can be computed in parallel for all $t$ using the \texttt{cumsum} operation over the matrix of gradients $[\nabla_{b} \mathcal{L}_{1}, ..., \nabla_{b} \mathcal{L}_{T}]$, which can be obtained using standard backpropagation and stored in memory. % because its space requirement is the same as the set of activations.

%\paragraph{Generation.}
The parallelized FWL updates are not usable during generation, which inherently has to produce one token at a time. 
Instead the model at each step (1) samples $\hat{x}_t$ from its output distribution using the current weights $\theta'$, (2) computes the gradient $\nabla_{\theta} \hat{\mathcal{L}_t}$ of the slow weights predicting $\hat{x}_t$, and (3) updates the fast weights as $\theta' \leftarrow \theta' - \alpha \circ \nabla_{\theta} \hat{\mathcal{L}_t}$. 
This process is still relatively efficient because the additional backward/forward passes are only applied to the FWL, not the whole transformer.

FWL gradient updates may first seem strange in that the model is ``training" on test sequences during perplexity evaluation or on its own outputs during generation.
We would like to emphasize that this way of conditioning on previous tokens is just as valid as standard transformer LMs attending over gold-standard tokens (from teacher forcing) or their own generated tokens (during sampling). 
Indeed, (\ref{eqn:2}) shows that FWLs can be viewed as adding another attention layer to the model, but with a pre-specified function computing the values (the gradient) rather than a learned matrix-multiply. 
%While many text generation systems suffer from producing repetitive outputs \citep{Holtzman2020TheCC}, 
%While neural text generation systems sometimes produce repetitive outputs \citep{Holtzman2020TheCC},
Empirically, we did not find FWLs to produce degenerate repetitive outputs \citep{Holtzman2020TheCC} more than baselines. 

%The slow weights and step sizes $\alpha$ are learned end-to-end. %requires roughly 30\% more compute that standard training. Most of the overhead is due to additional matrix-multiplies with the embedding matrix. 
%second-order gradients. 

%\vspace{1mm}
\paragraph{Training.}
Training jointly optimizes the transformer, softmax, and FWL parameters, as well as the step sizes $\alpha$, to minimize the combined loss $\sum_{t=1}^T \mathcal{L}'_t$ over training sequences. 
We compute second-order gradients for the FWL parameters, backpropagating through the gradient updates $\gradt{\theta}$. Intuitively, this means that in addition to learning to {\it expect} the gradient update and adapt quickly, the model also learns to {\it produce} effective gradient updates. 
While second-order gradients are expensive to compute for some models, they are not for FWLs because there is no back-propagation through time for the FWL parameters. 

%Training requires computing second-order gradients, but this in practice does not add extra compute since there is no backpropagation through time. 

%\vspace{1mm}
\paragraph{Runtime.}
FWLs add relatively modest compute and memory overhead to the model.
%For sparse transformers, FWLs add
In our experiments, FWLs add $<$30\% overhead in FLOPs and $<$20\% in wall clock time to both sparse transformers and Transformer-XL when scoring text perplexity.
The majority of this extra compute comes from the two extra passes through the output layer (the backward pass and fast weight pass), although using an adaptive softmax somewhat alleviates this cost.
In contrast, dynamic evaluation adds over $200\%$ extra compute due to needing an extra backward and forward pass through the whole transformer. 

%\vspace{1mm}
\paragraph{Connection to Meta-Learning.}
FWLs can be viewed as applying gradient-based meta-learning to LMs. 
Specifically, language modeling is treated as a few-shot learning task where the support set contains the tokens seen so far and the query set contains the next token.
The FWL training is essentially using MAML \citep{finn2017model}, where there is a single inner loop optimization step over the support set that adapts the model parameters using a gradient update.
%The FWL training uses MAML \citep{finn2017model} with only one inner-loop optimization step over the support set: the fast weight gradient update. 
This connection helps explain a surprising property of FWLs: although it applies the fast weight gradient update to only a small subset of the network, it performs comparably to dynamic evaluation, which updates all transformer parameters.
\citet{raghu2019rapid} show that MAML works just as well when the inner-loop update is only applied to the last layer of the network, which is similar to how FWLs only update a few weights on top of the transformer. 
\paragraph{Connection to Fast Weight Programmers (FWPs).}
FWPs use a slow-weight neural network to generate fast weights for another net \citep{schmidhuber1992learning,irie2021going}. \citet{schlag2021linear} show that FWPs can be viewed as transformers with linear attention, similar to the linear attention FWLs employ.
However, FWLs use gradient information to provide the update rather than generating the update from a network.

%It may at first seem strange to ``train" on test sequences through the gradient update or 

%transformers with linear attention can be viewed as FWPs.% FWLs also pose the fast weight update as linear attention. 

% Linear attention methods \citep{wang2020linformer,hua2022transformer} are particularly relevant to our work because the fast weight update can be rephrased as linear attention. 

%Indeed, \citet{schlag2021linear} show that linear transformers with are essentially acting like fast weight networks, which is similar to the connection between FWL gradient updates and linear attention we discuss.

\section{Experiments}

\begin{table}[tb]
\begin{center}
\small
\begin{tabularx}{\linewidth}{X l l}
\ttop
\textbf{Method} & \textbf{PPL} & \textbf{Tok/s} \tstrut \tsep 
Compressive Transformer & 17.1 & -- \tstrut \\
Routing Transformer & 15.8 & -- \\
kNN-LM & 15.8 & -- \tsep 
Transformer XL & 18.1 & 1575 \tstrut \\
Transformer XL + Dynamic Eval & 16.4 & 510 \tsep
Transformer XL + FWL (ours) & 16.6 & 1340 \tstrut 
\tbottom
\end{tabularx} 
\end{center}
\vspace{-1mm}
\caption{Test set WikiText-103 perplexities and inference speeds (on one V100 GPU).}
\label{tab:main}
\end{table}

\begin{figure*}[tb!]
\begin{center}
\begin{minipage}[c]{0.345\textwidth}
\includegraphics[width=\textwidth]{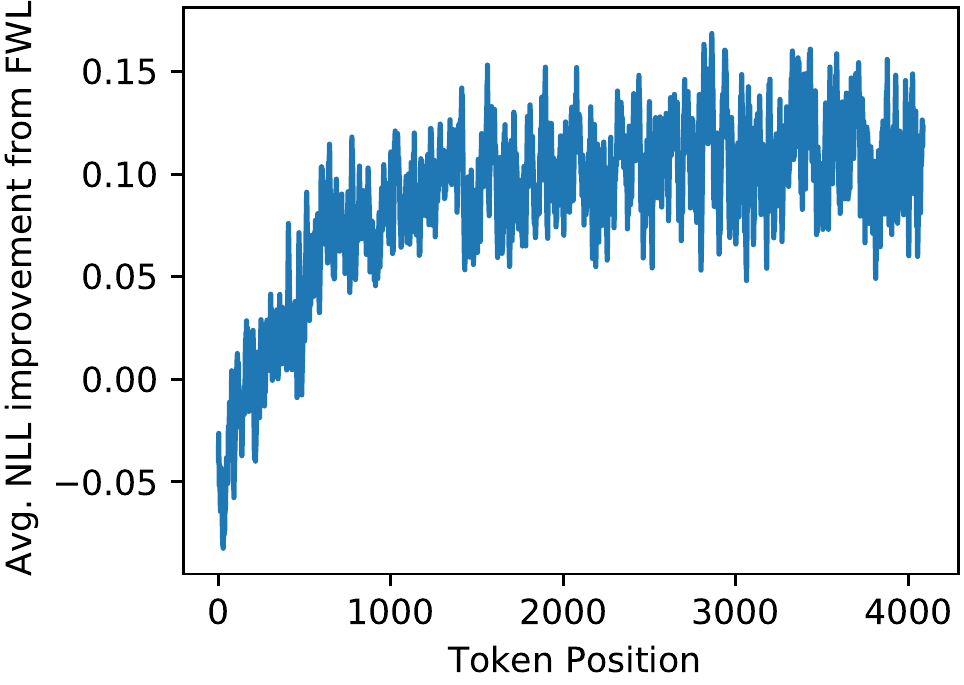}
\end{minipage}
\hspace{2.5mm}
\begin{minipage}[c]{0.43\textwidth}
\includegraphics[width=\textwidth]{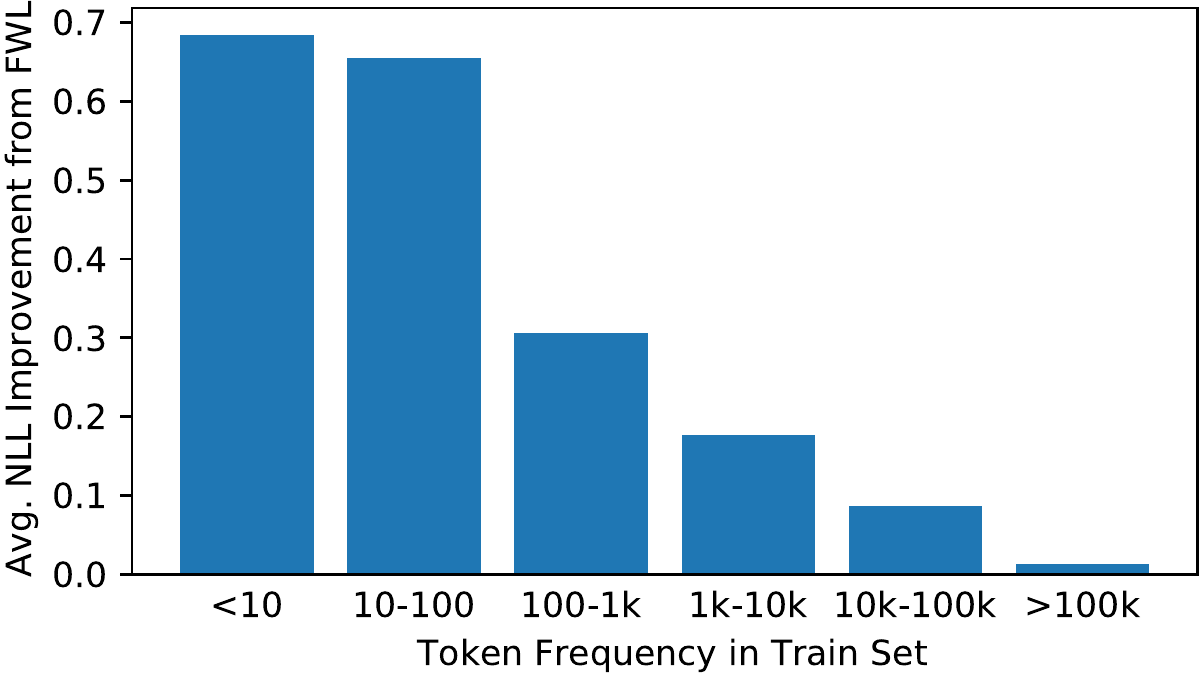}
\end{minipage}
\hspace{2.5mm}
\begin{minipage}[c]{0.16\textwidth}
\includegraphics[width=\textwidth]{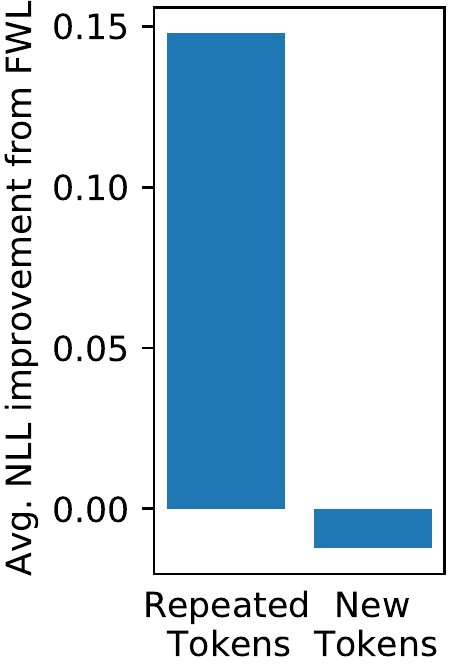}
\end{minipage}
\end{center}
\vspace{-1mm}
\caption{
Per-token negative-log-likelihood improvements over the baseline.
FWLs most improve LMs on long documents, rare tokens, and tokens repeated multiple times in the text. As the WikiText-103 dev set is small, we use a sparse transformer trained on 2/3 of the train set and evaluated on the other 1/3 to produce more robust results. }
\vspace{-1mm}
\label{fig:analysis}
\end{figure*}

% \subsection{Setup}

%\xhdr{Dataset.}
We experiment on WikiText-103 \citep{merity2016pointer}, a standard benchmark for language modeling consisting of approximately 100M tokens from English Wikipedia. We report perplexity using the standard tokenization and splits. 
%\vspace{1mm}
%\xhdr{Models.}
We consider two baseline models: the sparse local attention transformer from \citet{roy2021efficient} and Transformer-XL \citep{dai2019transformer}. % a transformer with added recurrence to process long sequences. 
The sparse transformer model has 12 layers with 768 hidden units (121M parameters). 
For Transformer-XL, we use the large model (257M parameters).

Transformer-XL processes the text recurrently: at each step it trains on one chunk while attending over but not backpropagating into a previous chunk.\footnote{Unfortunately, this recurrence means we lose the parallelism advantage of FWLs, although the other benefits over dynamic evaluation remain.} 
We use an analogous trick with FWLs to efficiently use them with the recurrent model: we keep a running fast weight update $\Delta_\theta$ from the tokens in previous chunks and similarly don't backpropagate into this update. 
More specifically, when processing a chunk of text $[x_S, x_{S + 1}, ..., x_T]$ the fast weight output $o_t$ is
\alns{
    o'_t = v'_t (W - \alpha_{W} \Delta_{W} - \alpha_{W}&\sum\nolimits_{i=S}^t \gradi{W}) \\
    \Delta_{W} \leftarrow \gamma_{W} \Delta_{W} + \text{stopgrad}(&\sum\nolimits_{i=S}^T \gradi{W})
}
where $\gamma_{W}$ is a learned decay factor. To reduce the training cost, we add the FWLs on top of the publicly released pre-trained Transformer-XL model and then fine-tune for 20K steps rather than training it from scratch.

%We keep a running fast weight update $\nabla_{\theta} \mathcal{L}_{1:}$

%We keep a running summed gradient 

%\vspace{-1mm}
\paragraph{Main Results.}
Table~\ref{tab:main} shows the results on the WikiText-103 test set. FWLs improve Transformer-XL by 1.5 perplexity points. This gain is comparable to the improvement from dynamic evaluation obtained by \citet{krause2019dynamic}, but FWLs are about 3x faster to run.
While recent methods such as Routing Transformer \citep{roy2021efficient} and KNN-LM \citep{khandelwal2019generalization} achieve better perplexity, they are not directly comparable because they use different base models; we expect FWLs could also be combined with them to improve results. 

%could be added to these strong models

%\vspace{-1mm}
\paragraph{Ablations.}
We compare different variants of FWLs in Table~\ref{tab:abl}. 
First we consider only training the slow weights of the FWL (i.e., using $\mathcal{L}_t$ instead of $\mathcal{L}'_t$), but then applying the fast weight update at test time.\footnote{We use a global step size that is tuned on the dev set.}
This method essentially applies dynamic evaluation to only a few layers of the network. 
While not performing as well as full dynamic evaluation, it still provides a sizable improvement given the small number of updated parameters and much faster inference speed.
The remaining gap to dynamic evaluation is closed when the FWL is used during training so that the model learns to benefit from the previous gradients. 

One hypothesis for the benefit of FWLs is that it biases the model towards copying seen tokens. 
To test this, we train a much simpler version of FWLs where only the bias term of the output softmax has fast weights applied.
We found this did not significantly improve results, perhaps because attention is sufficient for the model to do this kind of copying, suggesting that FWLs are learning more complicated updates than just biasing the model towards repeating seen tokens.

\begin{table}[tb]
\begin{center}
\small
\begin{tabularx}{\linewidth}{X l l}
\ttop
\textbf{Method} & \textbf{PPL Sparse} & \textbf{PPL XL} \tstrut \tsep 
No FWLs & 25.1 & 17.3 \tstrut \\
FWLs & 22.4 & 15.9 \\
Test-time only & 24.1 & 16.7 \\
%No second order grads & &  \\
%3 Gradient updates & & \\
Bias Only & 25.0 & 17.3 \\
Dynamic Evaluation & 22.4 & 15.8
\tbottom
\end{tabularx} 
\end{center}
\vspace{-1mm}
\caption{Dev set WikiText-103 perplexities for various ablations on \textbf{Sparse} transformer and Transformer-\textbf{XL}.}
\label{tab:abl}
\end{table}

\paragraph{Where do FWLs help?} Figure~\ref{fig:analysis} shows which tokens are better predicted by a FWL-augmented model. First, we find improvements are larger for tokens toward the end of input, implying that FWLs help models make use of long contexts and work best on long documents. Intuitively, more previous tokens will provide a better gradient estimate, similar to how meta-learning methods benefit for a larger support set. 
Next, we find FWLs help most on rare tokens, perhaps because they require better modeling of long contexts to predict.
Lastly, we find FWLs actually make the model slightly worse at predicting a token the first time it appears in a text, but help when the token has occurred previously (a net gain because around 70\% of tokens in WikiText-103 are repeats).

\section{Conclusion}
Fast Weight Layers provide the benefits of dynamic evaluation at a fraction of the compute cost and memory usage. They can easily be added to existing language models and yield strong results on language modeling benchmarks.
% , for example performing competitively with kNN-LM \citep{khandelwal2019generalization}, which uses an external memory containing 100M tokens.
Applying FWLs to few-shot learning tasks is one interesting future direction:
doing one (or perhaps a small number) of gradient updates on few-shot examples might offer a nice middle ground in-context learning where the model parameters are fixed and full fine-tuning. Indeed, \citet{yoshida2021reconsidering} show that hidden state optimization, a method closely related to dynamic evaluation, can improve few-shot LM performance. 

%Gradient-based fast weight methods such as FWLs may offer a middle ground between full fine-tuning and in-context learning with language models, which would be an interesting direction to explore in future work.

\section{Limitations}
FWLs can be viewed as an inductive bias encouraging the model to adapt to previous tokens.
As an inductive bias, their value may be limited for larger models trained on larger datasets. 
While our experiments show FWLs improve models with hundreds of millions of parameters, initial experiments with bigger models suggest that their benefit decreases as models get larger, and we think it is unlikely that an add-on like a FWL will substantially improve models of the scale of \mbox{GPT-3} \citep{brown2020language}.
Furthermore, we have shown that using FWLs at training time makes them more effective, but this has a disadvantage as well. FWLs can't be directly applied to already-trained transformer language models the way dynamic evaluation can: some fine-tuning with the fast weight layer added is required. Lastly, while we have shown FWLs improve LM perplexity, we have not evaluated FWLs at other text generation tasks, which we leave for future work.

\section*{Acknowledgements}
We thank Urvashi Khandelwal and the anonymous reviewers for their thoughtful comments and suggestions. 
%As an inductive bias in the model, the benefit of %FWLs may not be as big for large language models. 

\bibliographystyle{acl_natbib}
\bibliography{emnlp2022.bib}

\end{document}